\documentclass[runningheads]{llncs}
\usepackage{graphicx}
\usepackage{tikz}
\usepackage{comment}
\usepackage{amsmath,amssymb} % define this before the line numbering.
\usepackage{color}

\usepackage[accsupp]{axessibility}  % Improves PDF readability for those with disabilities.

\usepackage{wrapfig}
\usepackage{multirow}
\usepackage{booktabs}
\usepackage{soul}
\usepackage{CJK}

\begin{document}
% \renewcommand\thelinenumber{\color[rgb]{0.2,0.5,0.8}\normalfont\sffamily\scriptsize\arabic{linenumber}\color[rgb]{0,0,0}}
% \renewcommand\makeLineNumber {\hss\thelinenumber\ \hspace{6mm} \rlap{\hskip\textwidth\ \hspace{6.5mm}\thelinenumber}}
% \linenumbers
\pagestyle{headings}
\mainmatter
\def\ECCVSubNumber{0}  % Insert your submission number here

\title{Boosting Multi-Modal E-commerce Attribute Value Extraction via Unified Learning Scheme and Dynamic Range Minimization} % Replace with your title

% INITIAL SUBMISSION 
%\begin{comment}
%\titlerunning{ECCV-22 submission ID \ECCVSubNumber} 
%\authorrunning{ECCV-22 submission ID \ECCVSubNumber} 
%\author{Anonymous ECCV submission}
%\institute{Paper ID \ECCVSubNumber}
%\end{comment}
%******************

% CAMERA READY SUBMISSION
%\begin{comment}
\titlerunning{Boosting MEAVE via ULS-DRAM}
% If the paper title is too long for the running head, you can set
% an abbreviated paper title here
%
\author{Mengyin Liu\inst{1} \and
	Chao Zhu\inst{1}\thanks{Corresponding author} \and Hongyu Gao\inst{1} \and \\
	Weibo Gu\inst{2} \and Hongfa Wang\inst{2} \and Wei Liu \inst{2} \and Xu-cheng Yin \inst{1}
}
\authorrunning{Liu et al.}
% First names are abbreviated in the running head.
% If there are more than two authors, 'et al.' is used.
%
\institute{
School of Computer and Communication Engineering, \\ University of Science and Technology Beijing, Beijing, China \\ \email{blean@live.cn, \{chaozhu,xuchengyin\}@ustb.edu.cn, \\ hanleygao0520@gmail.com} \\
\and Data Platform Department, Tencent, Shenzhen, China \email { \{lukegu,hongfawang\}@tencent.com, wl2223@columbia.edu}
}

%\institute{Princeton University, Princeton NJ 08544, USA \and
%	Springer Heidelberg, Tiergartenstr. 17, 69121 Heidelberg, Germany
%	\email{lncs@springer.com}\\
%	\url{http://www.springer.com/gp/computer-science/lncs} \and
%	ABC Institute, Rupert-Karls-University Heidelberg, Heidelberg, Germany\\
%	\email{\{abc,lncs\}@uni-heidelberg.de}}

%\end{comment}
%******************
\maketitle

\begin{abstract}
   With the prosperity of e-commerce industry, various modalities, e.g., vision and language, are utilized to describe product items. It is an enormous challenge to understand such diversified data, especially via extracting the attribute-value pairs in text sequences with the aid of helpful image regions. Although a series of previous works have been dedicated to this task, there remain seldomly investigated obstacles that hinder further improvements: 1) Parameters from up-stream single-modal pretraining are inadequately applied, without proper jointly fine-tuning in a down-stream multi-modal task. 2) To select descriptive parts of images, a simple late fusion is widely applied, regardless of priori knowledge that language-related information should be encoded into a common linguistic embedding space by stronger encoders. 3) Due to diversity across products, their attribute sets tend to vary greatly, but current approaches predict with an unnecessary maximal range and lead to more potential false positives. To address these issues, we propose in this paper a novel approach to boost multi-modal e-commerce attribute value extraction via unified learning scheme and dynamic range minimization: 1) Firstly, a unified scheme is designed to jointly train a multi-modal task with pretrained single-modal parameters. 2) Secondly, a text-guided  information range minimization method is proposed to adaptively encode descriptive parts of each modality into an identical space with a powerful pretrained linguistic model. 3) Moreover, a prototype-guided attribute range minimization method is proposed to first determine the proper attribute set of the current product, and then select prototypes to guide the prediction of the chosen attributes. Experiments on the popular multi-modal e-commerce benchmarks show that our approach achieves superior performance over the other state-of-the-art techniques.

\end{abstract}

\section{Introduction}

\begin{figure}[t]
	\centering
	\includegraphics[width=1\textwidth]{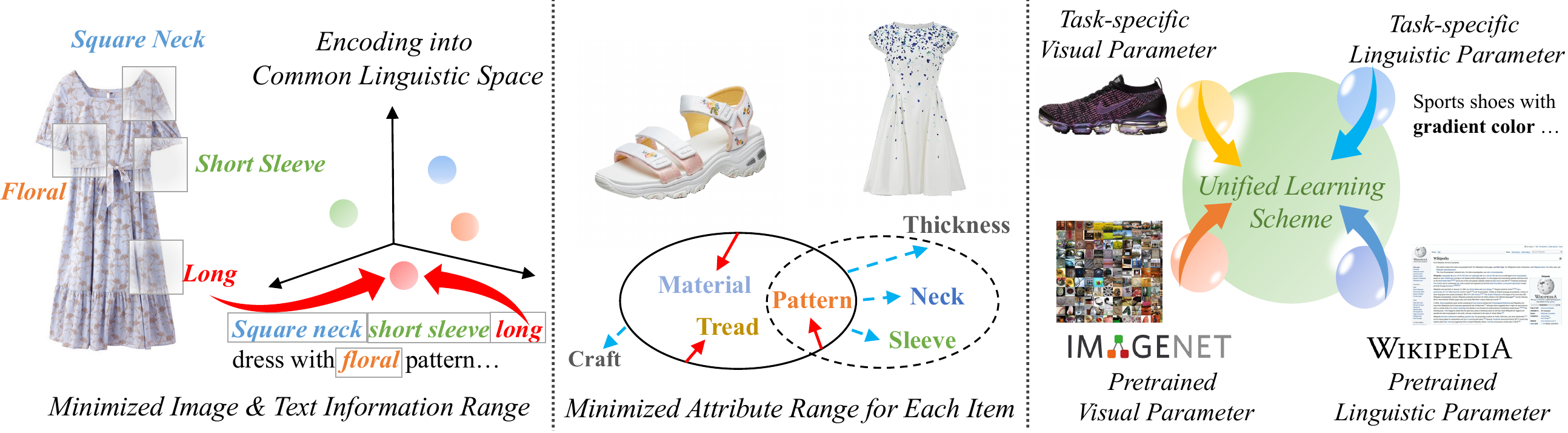}
	\caption{
		Proposed Dynamic Information Range Minimization, Dynamic Attribute Range Minimization, and Unified Learning Scheme.  \textbf{Left: }descriptive parts of image and text information are minimized in gray boxes and then encoded into a common linguistic space via a strong encoder. \textbf{Middle: }due to the diversity among products, attribute range of each item is minimized in solid ovals for more accurate value extraction. \textbf{Right: }common knowledge from up-stream pretraining and task-specific one from down-stream learning in colorful balls are combined via unified learning. Best viewed in colors.
	}
	\label{figure1}
\end{figure}

Understanding a massive amount of product data plays a significant role in e-commerce applications, including recommendation  \cite{xian2021ex3,hou2019explainable} and question answering \cite{gao2019product}. Among diverse strategies, attribute value extraction is capable of formatting unstructured textual descriptions into structured attribute-value pairs, which highlights the most interesting characteristics of products. 

Specifically, in a sequence tagging manner, attribute values are directly extracted from the original text input, via the prediction with identical length in a certain symbol set, e.g., ``B-I-O'' marks the begin, inside, and outside parts (``not-a-value'') of attribute values. Furthermore, with the progress of marketing methods in e-commerce scenarios, images, as one of the major visual modalities, have been introduced to more precisely describe product characters. ``Blue'', as an example, means either emotional or color attribute in plain texts, but images of the product help customers judge intuitively. Thus, it is natural for an extractor to take full advantage of vision to obtain attribute values.

Although some recent multi-modal extractors adopt both vision and language inputs \cite{zhu2020multimodal,logan2017multimodal,lin2021pam}, they make insufficient use of single-modal knowledge from large-scale pretraining. For instance, JAVE \cite{zhu2020multimodal} and MAE-model \cite{logan2017multimodal} rely on feature representations of input data, which are generated by pretrained visual \cite{he2016deep} and linguistic \cite{devlin2018bert} encoders with fixed parameters.  Most of them do not explain for such design, but we think it might be a conventional belief that fixed pretrained models are capable of handling most downstream tasks. In fact, it is hard for them to cover all corner cases in e-commerce, e.g., fashion style in 2020s hardly takes place in ImageNet dataset \cite{deng2009imagenet} collected in around 2009. Hence, as shown in right part of Figure \ref{figure1}, a unified learning scheme is proposed to bridge the gap between up-stream pretraining and down-stream tasks.

Meanwhile, during perceiving multi-modal information of a human customer, visual elements are converted into semantic symbols, e.g., colors, shapes, or sizes, to make a decision, similarly for a multi-modal attribute extractor. More precisely, it is priori knowledge that non-textual descriptive clues from other modalities should be encoded with textual information into a common feature space, as semantic symbols like those in human minds. However, previous methods employ either naïve fusion or simple gate mechanisms, which is powerless to dynamically model such a language-dominated space by newly initialized and deficient parameters, compared with strong pretrained linguistic encoders illustrated in the left part of Figure \ref{figure1}.

Besides, diversity of products, e.g., taxonomies or multi-functional purposes, leads to distinct ranges of attributes among them. Conversely, most attribute value extractors consider the attribute range as a maximal union set across all the products. In details, ``B'' and ``I'' symbols in sequence tagging outputs are appended with attribute names, such as ``B-battery\_life'', ``I-sleeve\_length''. Quantitatively, every textual position is forced to be classified into twice of all attribute amount with an ``O'', denoted as $T_o = (C\times| \{B, I\} | + | {O} | ) \times S$, where $C$ is the attribute amount and $S$ is text length. This extreme inconsistency between data distribution and modeling leads to non-essential classification outputs, which brings about more potential false positives and thus leads to a lower generalization capacity.

\begin{figure}[t]
	\centering
	\includegraphics[width=1\textwidth]{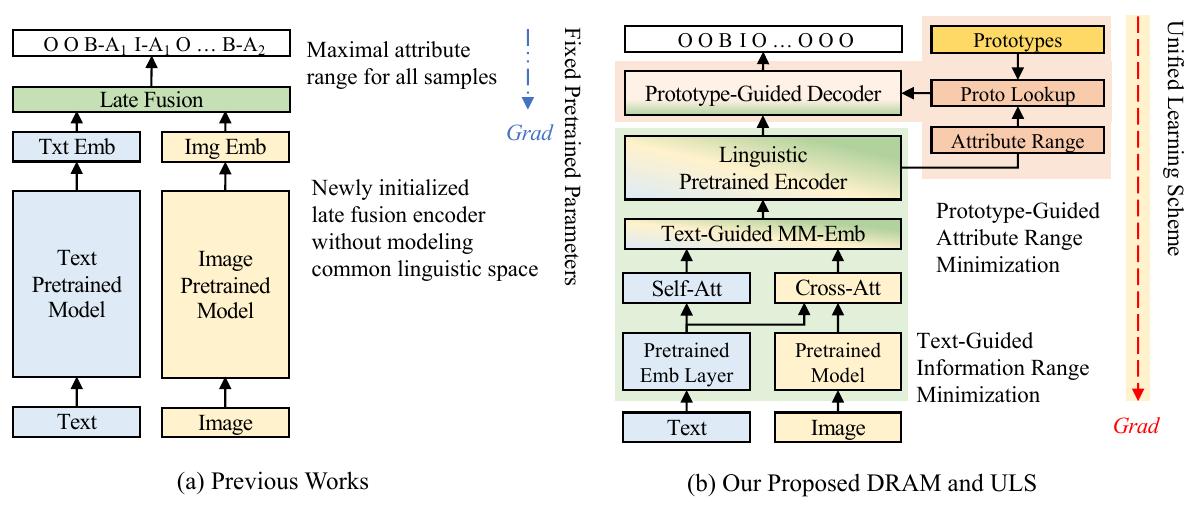}
	\caption{Comparison of pipelines between previous works and our proposed Dynamic Range Minimization (DRAM) and Unified Learning Scheme (ULS, marked near the longer top-down arrow).``\textit{Grad}'' means gradient propagation.}
	
	\label{figure2}
\end{figure}

Differently, we redesign a two-stage approach to minimize the attribute range of specific product dynamically as shown in middle of Figure \ref{figure1},  which has not been investigated before. Firstly, it determines the attribute range of the current sample, via only once multi-label classification.  Secondly, it selects learnable prototypes of chosen attributes. Finally, only ``B-I-O'' sequences are predicted for only chosen attributes under the conditional guidance of their prototypes, where the number of classification is denoted as $T_m = C + (| \{B, I, O\} | \times S) \times C_m$, where $C_m$ is the number of selected attributes ($C_m << C $, thus $T_m << T_o$).

In conclusion, we have observed that: the problems of unified learning and range minimization are remained to be investigated for boosting the extraction of attribute-value pairs on multi-modal e-commerce data. Therefore, we propose a novel attribute value extraction approach to jointly learn the pretrained as well as task-specific parameters and to dynamically minimize the ranges in both modality and attribute perspectives. The main contributions of this paper are summarized as follows:

\begin{itemize}
	\item We propose a \textbf{U}nified \textbf{L}earning \textbf{S}cheme (\textbf{ULS}) for an efficient combination of common knowledge from up-stream pretraining and task-specific knowledge from down-stream multi-modal attribute value extraction learning, by jointly training the corresponding parameters in an end-to-end style.
	\item We propose a novel \textbf{D}ynamic \textbf{RA}nge \textbf{M}inimization approach, named \textbf{DRAM}. Following the intuition that only language-related parts of each modality contribute to the extraction of textual values, Text-Guided Information Range Minimization (TIR) dynamically extracts the descriptive parts of both text and image. Moreover, it encodes all the informative clues into a common embedding space with a powerful pretrained linguistic model.
	\item As another component of DRAM, Prototype-Guided Attribute Range Minimization (PAR) is designed in a two-stage manner to dynamically determine the range of attributes. Within the minimized range, learnable prototypes are selected to guide value extraction of the chosen attributes. Complementing TIR and PAR, our proposed \textbf{ULS-DRAM} achieves state-of-the-art performance on popular MEPAVE and MAE benchmarks. 
\end{itemize}

\section{Related Works}

\subsection{Single-Modal Attribute Value Extraction}

Based on the text-centric task characteristic, early works of attribute value extraction are designed to accept single-modal textual inputs. 

For instance, OpenTag \cite{zheng2018opentag} adopts a basic pipeline of sequence tagging with a multi-class prediction of sequences. Hence, each position should be classified into around twice of all attribute amount, leading to the failures on large-scale datasets with thousands of attributes. Similarly, slot-filling task also performs multi-task sequence tagging for extracting values. RNN-LSTM \cite{hakkani2016multi} and Attn-BiRNN \cite{liu16c_interspeech}  adopt pretrained word embeddings and Recursive Neural Networks. Slot-Gated \cite{goo2018slot} further introduces a gate mechanism for a more adaptive model. The Pretrained BERT \cite{devlin2018bert} model empowers Joint-BERT \cite{chen2019bert} via jointly fine-tuning on down-stream task. Recently, K-PLUG \cite{xu2021k} collects 25M extra data to pretrain text encoder and also fine-tunes it on down-stream tasks. Instead, our method fine-tunes not only linguistic \cite{devlin2018bert,liu2019roberta} but also visual pretrained models \cite{he2016deep} for better performance without any extra data.

\setlength\intextsep{0pt}
\begin{wrapfigure}[18]{r}{16.5em}
	\includegraphics[width=1.0\linewidth]{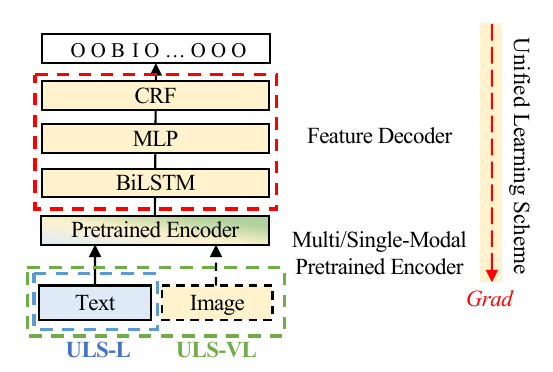}
	\caption{The basic model trained with ULS. The encoder is initialized with the pretrained weights. Identified by input modality, ULS-L denotes the text-only version and ULS-VL the multi-modal one, respectively. ``\textit{Grad}'' means gradient propagation.}
	\label{figure3}
\end{wrapfigure}

Based on the architecture of OpenTag, SUOpenTag \cite{xu2019scaling} employs a novel prediction style with given attributes, via encoding current attributes and input texts with pretrained BERT \cite{devlin2018bert} and calculating the co-attention between them, thus only one sequence of one attribute should be predicted and fits on large-scale datasets. Following this prototype, AVEQA \cite{wang2020learning} further introduces distillation training from a large pretrained model and an auxiliary matching task between text and attribute; AdaTag \cite{yan2021adatag} converts attribute encoding into a dynamic network architecture. However, their predictions are limited by known attribute ranges of input texts. If the range is unavailable in the inference time, maximum range with all the attributes is utilized and leads to more wrong predictions. 

Our proposed approach tackles such a limitation by dynamically determining the minimal attribute range of current inputs, via an extra multi-label classification task on attributes. 

\subsection{Multi-Modal Attribute Value Extraction}

In the meantime, attribute value extractors can also benefit from the assistance of more modalities to make prediction, especially visual images with rich clues such as colors, shapes, or sizes. 

MAE-model \cite{logan2017multimodal} constructs a generative baseline with inputs of image, text, and attribute. Single-modal features are extracted by fixed pretrained parameters, and then concatenated as a late fusion to directly predict the specific value via classification. Extra modality of Optical Character Recognition (OCR) empowers PAM \cite{lin2021pam} to perceive information on images more efficiently and all the modalities are processed by a unified Transformer architecture to generate results. Whereas the stability of prediction is hard to ensure without a constraint from input texts. 

On the contrary, as a sequence tagging approach, JAVE \cite{zhu2020multimodal}  jointly predicts attributes as well as values based on image and text inputs. Nevertheless, it also bears a series of shortcomings: 1) Parameters from single-modal pretraining are fixed as \cite{logan2017multimodal}, which is insufficient for the adaption to a multi-modal down-stream task. 2) It is hard for newly initialized late-fusion gate networks in JAVE to model a common space for both textual and non-textual information. 3) Range of attributes is not minimized by JAVE which makes multi-class sequence prediction, thus shares the similar problem with OpenTag \cite{zheng2018opentag} on datasets with a large attribute range.

To tackle the problems above, our proposed ULS performs unified learning on multi-modal tasks with up-stream pretrained and down-stream task-specific parameters. In the meantime, our proposed DRAM not only dynamically minimizes the information range via extracting the descriptive parts of each modality and encoding them into a common embedding space with a powerful pretrained linguistic model \cite{devlin2018bert,liu2019roberta}, but also minimizes the attribute range via first classification and then learnable prototype guidance.

\setlength\intextsep{10pt}
\begin{wrapfigure}[25]{l}{15.5em}
	\centering
	\includegraphics[width=1.0\linewidth]{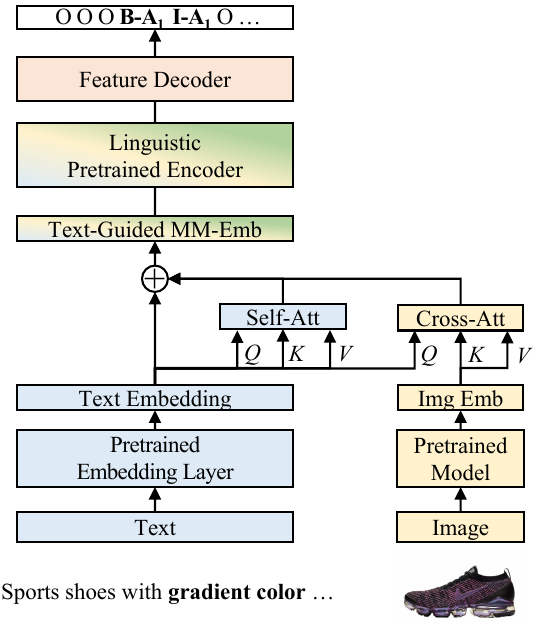}
	\caption{The overall architecture of our proposed Text-Guided Information Range Minimization (TIR).  As a component of DRAM, it extracts language-related parts of each modality via textual embeddings as queries. Then it encodes them into a common space by a strong linguistic pretrained encoder.}
	\label{figure4}
\end{wrapfigure}

\subsection{Multi-Modal Space via Linguistic Encoders}

Determined by the priori knowledge of specific tasks, pretrained linguistic encoders \cite{devlin2018bert,liu2019roberta} are applied to encode the multi-modal embeddings and model the common space, because they are talent to extract language-related discriminative representations. In Named Entity Recognition (NER), RpBERT \cite{sun2021rpbert} extracts textual named entities based on text and image, which considers language-related information to be significant. Hence,  it is initialized with pretrained BERT \cite{devlin2018bert} weights and fine-tuned with multi-modal inputs. Similarly, to learn unsupervised multi-modal representations, a series of models also adopt a pretrained linguistic encoder, e.g., VL-BERT \cite{su2019vl} and UNITER \cite{chen2020uniter}. Detected objects or image regions are processed as special text embeddings together with text inputs, thus a common space with latent semantic representation is constructed. 

Differently, our proposed TIR not only models a unified multi-modal space via powerful linguistic encoders \cite{devlin2018bert,liu2019roberta}, but also extracts language-related parts of each modality to minimize the information range for a space without redundant distractive information. 

\section{Proposed Method}

As illustrated in Figure \ref{figure2}(b), our proposed method ULS-DRAM comprises two major components: Unified Learning Scheme (ULS) and Dynamic Range Minimization (DRAM). Furthermore, DRAM can be divided in two-folds: Text-Guided Information Range Minimization (TIR) and Prototype-Guided Attribute Range Minimization (PAR). ULS efficiently utilizes the common knowledge from up-stream single-modal pretraining and task-specific one from down-stream multi-modal learning, via jointly training the pretrained parameters and down-stream task ones. 

As the essential parts of DRAM, TIR is designed to dynamically extract the descriptive parts in a minimal range of both text and image and then encode them into a common embedding space with a powerful pretrained linguistic model. PAR is designed to adaptively determine the minimized range of attributes for the current input and then guide value extraction for the selected attributes via learnable prototypes. ULS and DRAM cooperate to boost the performance of attribute value extraction. More details will be introduced in the following sections.

\begin{figure}[t]
	\centering
	\includegraphics[width=1.0\textwidth]{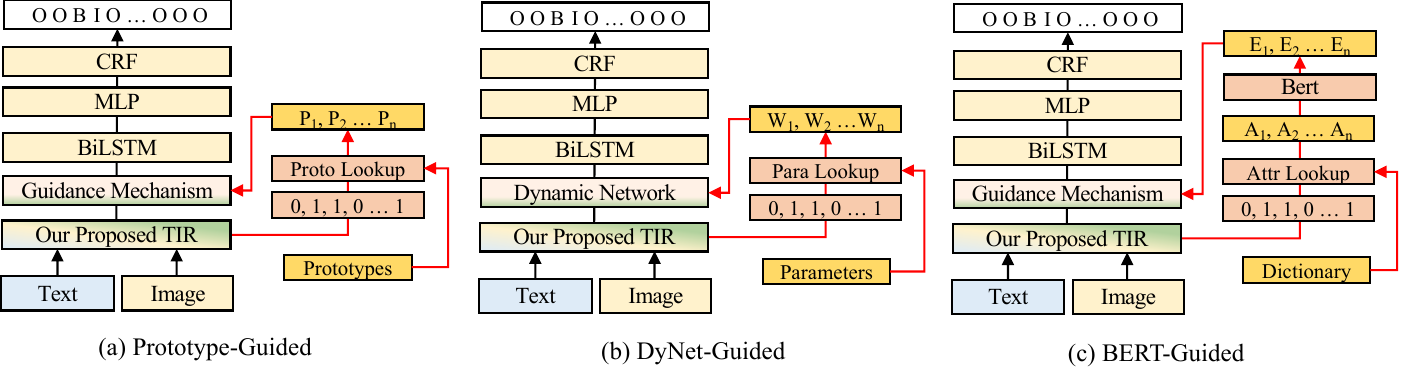}
	\caption{Different variants of the proposed Attribute Range Minimization: (a) Prototype Guided Policy learns ``Prototype'' embeddings for each attribute, and then guides the Feature Decoder to predict values of attributes in minimized range with the Guidance Mechanism. (b) DyNet Guided Policy learns parameters instead and applies them on the Dynamic Network. (c) BERT Guided Policy encodes selected attribute words from a  dictionary with the pretrained BERT model into textual embeddings for guidance.}
	\label{figure5}
\end{figure}

\subsection{Unified Learning Scheme}

Considering differences between up-stream pretraining and down-stream task training, we find there is a significant gap in the learned knowledge from them. 

For example, in the field of linguistic pretraining, popular pretrained model BERT \cite{devlin2018bert} adopts self-supervised learning on a large-scale unannotated word corpus from websites such as Wikipedia. While as to visual pretraining, commonly used ResNet \cite{he2016deep} series are trained in a supervised scheme on the ImageNet dataset \cite{deng2009imagenet} with human-crafted class annotation for each Internet collected image. More differently, the down-stream task in this paper is multi-modal attribute value extraction in e-commerce scenarios, where most data is related to product descriptions with both texts and images, other than wiki articles or natural images. 

Such a giant data distribution inconsistency reminds us of the fact that various previous works \cite{zhu2020multimodal,logan2017multimodal,hakkani2016multi,liu16c_interspeech,goo2018slot} only fix the pretrained parameters to extract feature embeddings and lead to a potential performance loss. Therefore, we re-design the training scheme as a unified one, named Unified Learning Scheme (ULS), which jointly fine-tunes pretrained single-modal parameters and task-specific multi-modal ones. 

In Figure \ref{figure3}, different basic model versions are designed for text-only single-modal and multi-modal unified learning, denoted as ULS-L and ULS-VL. For text inputs, a pretrained embedding layer \cite{devlin2018bert} converts the tokenized text IDs into high-dimension embeddings. As for images, a pretrained ResNet \cite{he2016deep} model is jointly fine-tuned and generates $7\times7$ patches of image embeddings, which are flattened into the length $49$ and linearly projected to the same dimension of text embeddings. For ULS-VL, two sequences are finally concatenated into one. Then ``Feature Decoder'', i.e., BiLSTM+CRF \cite{zheng2018opentag}, decodes features into predictions.

Hence, the gap mentioned above is eliminated and an efficient combination by ULS, which bridges between common knowledge from up-stream pretraining and task-specific knowledge from down-stream multi-modal attribute value extraction learning, leads to a better generalization capability and performance.

\subsection{Text-Guided Information Range Minimization}

When browsing the e-commerce web pages, there are multiple information sources for a human customer to perceive, including visual and textual ones. During the processing in the brain, visual elements are converted into semantic conceptual symbols, e.g., colors, shapes, and sizes, just as keywords from texts. Based on the explicit symbols from visual and textual elements, a customer can make a further decision, similarly for an attribute value extractor.

Following the priori knowledge above, we consider that descriptive clues from each modality should be encoded into \textit{a common linguistic space}, as semantic symbols like those in human minds. For an extractor, information range of each modality should be minimized to only language-related parts to describe products. Meanwhile, such a space is difficult for newly initialized networks to model without common knowledge from large-scale linguistic pretraining. 

Hence, we design a novel Text-Guided Information Range Minimization (TIR) approach to implement these discussed ideas. As an efficient feature extractor for both intra and inter information, the attention mechanism \cite{shaw2018self} is the best choice for constructing our pipeline as an essential component. Given two input sequences $x_Q \in \mathbb{R}^{S_{1} \times D}$, $x_K = x_V \in \mathbb{R}^{S_{2}\times D}$, they are linearly projected by weight matrices $W_Q$, $W_K$, and $W_V \in \mathbb{R}^{D \times D'}$, where $S_1$ and $S_2$ denote sequence lengths of two inputs, $D$ and $D’$ denote the embedding dimensions and the internal ones of the attention mechanism. Thus, the linearly projected sequences are computed as:
\begin{equation}
	Q=x_QW_Q, K=x_KW_K, V=x_VW_V.
	\label{eq1}
\end{equation}

$Q, \in \mathbb{R}^{S_1 \times D’},$ $K$ and $V \in \mathbb{R}^{S_2 \times D’}$ represent three kinds of features, where $Q$ is a ``Query'' to retrieve useful parts of $V$ ``Value''. However, since $Q$ and $V$ might originate differently, e.g., they are either text or image, directly computing their matching degree is inappropriate. $K$ as another representation of each position in $V$ performs the ``Key'' role to match with $Q$. Thus, we have the formula of attention:
\begin{equation}
	Attention(K,Q,V)=\frac{QK^{\rm T}}{\sqrt{D'}}V.
	\label{eq2}
\end{equation}

Obtained via the dot product between $Q$ and transposed $K$, attention mapping $QK^T \in \mathbb{R}^{S_1 \times S_2}$ means the matching degree between each position of $Q$ and $K$, normalized by dimension numbers. Based on this dense mapping, all the positions of $V$ are multiplied with the corresponding matching score and summed for each position of $Q$. The result of Attention $A\in \mathbb{R}^{S_1\times D}$ has the same shape of $Q$, which represents the meaningful information in $V$ extracted for $Q$. 

In practice, due to the text-guided priori knowledge discussed before, text embeddings $x^t$ perform as $x_Q$ for not only self-attention inside the text modality, but also cross-attention between text and image modalities. For intra-modality self-attention, $x_K=x_V=x^t$. For inter-modality cross-attention, image embedding sequence $x_K=x_V=x^v$. Two attentions are formulated as:
\begin{equation}
	SelfAtt(x^t)=\frac{(x^tW_Q^t){(x^tW_K^t)}^{\rm T}}{\sqrt{D'}}x^tW_V^t,
	\label{eq3}
\end{equation}
\begin{equation}
	CrossAtt(x^t,x^v)=\frac{(x^tW_Q^v){(x^vW_K^v)}^{\rm T}}{\sqrt{D'}}x^vW_V^v.
	\label{eq4}
\end{equation}

As shown in Figure \ref{figure4}, with the self-attention and cross-attention, text embeddings play the roles as queries to guide the extraction of language-related information from both text and image embeddings from pretrained embedding layers or models. 
\begin{equation}
	E_m=x^t+(SelfAtt(x^t)+CrossAtt(x^t,x^v))W^m.
	\label{eq5}
\end{equation}

Finally, highlighted via TIR, descriptive parts share ``relevant'' meanings with parts of ``Query'' text.  They helps model to know which tokens of $x^t$ are supported by more multi-modal information via fusion. Hence, they are added together with $x^t$, like the \textit{positional/token-type embeddings} in BERT \cite{devlin2018bert} or ViT \cite{dosovitskiy2020image}. $E_m$ is obtained by Eq. \ref{eq5} and then encoded into a common semantic space by a strong linguistic encoder \cite{devlin2018bert,liu2019roberta}, where $W_m \in \mathbb{R}^{D\times D’}$ denotes re-projecting the dimension number of attention back to that of embedding.

\subsection{Attribute Range Minimization}

In e-commerce scenarios, difference among products varies in a large scale due to taxonomies or functions, etc. Therefore, attribute ranges of each product tend to be specialized, e.g., ``battery\_life'' attribute appears more frequently in electronic products than clothing or shoes. On the contrary, many previous works \cite{zhu2020multimodal,zheng2018opentag,hakkani2016multi,liu16c_interspeech,goo2018slot,chen2019bert} apply multi-class sequence prediction with a maximum union set across all products, leading to \textit{unnecessary false positives}. While the others predict for given attributes, there are still limitations when attribute sets of the current products are unavailable during the inference time. 

Having observed the inconsistency between data distribution and model design above, we propose a novel dynamic attribute range minimization approach to tackle it. Firstly, the model can predict the attribute range dynamically for the current sample. And the prediction can be formalized as a multi-label classification task:
\begin{equation}
	R_i=\sigma(f(T_i,\ V_i\ |\ \theta_{Pre},\ \theta_{TIR})W_{cls}).
	\label{eq6}
\end{equation}

$R_i$ denotes the attribute range of current sample $i$ with input text $T_i$ and image $V_i$. The multi-modal pretrained encoder shown in Figure \ref{figure3} is denoted as $f$ with pretrained parameters $\theta_{Pre}$ and down-stream task parameters for proposed TIR $\theta_{TIR}$ , which are jointly learned via proposed ULS-VL. $W_{cls}\in \mathbb{R}^{D\times C}$ projects the encoded multi-modal feature into attribute classes $C$, and a sigmoid function $\sigma$ is used to non-linearly normalize the class scores into multi-hot range digits $R_i\in [0, 1]$.

Secondly, attributes with higher score are obtained as the minimized range of the current input with a threshold $S_{thr}$. Given the dynamically determined minimized range of attributes, only tagging sequences of items within the range are predicted.  For a representation set $P$ of all attributes, the selected subset $P_{min}$ via the minimized range is: 
\begin{equation}
	P_{min}=P[R_i\ >\ S_{thr}].
	\label{eq7}
\end{equation}

As shown in Figure \ref{figure5}, to represent attributes and guide the model to predict attribute value within the minimized range, there are many feasible candidates. So three variants of policy are designed: (1) Prototype Guided Policy learns prototype embeddings of each attribute. (2) Instead, DyNet Guided Policy directly apply parameters corresponding to attributes within the minimized range into Dynamic Networks, thus without the need of guidance mechanism. (3) BERT Guided Policy employs a pretrained BERT \cite{devlin2018bert} model to explicitly encode selected attribute words from a dictionary into textual embeddings. (1) and (3) shares the same guidance mechanism.

Take (1) as an example. If an attribute ``Color'' is predicted inside the range by Eq. \ref{eq6}, its ``Prototype'', a jointly trained embedding vector, is selected by Eq. \ref{eq7}. The model is required to predict for ``Color'', only when its prototype is selected. Such a \textit{conditional guidance} is realized by our proposed mechanism in Eq. \ref{eq8}. It calculates cosine similarity scores of tokens and selected prototype. These scores indicate the existences of ``Color'' 's values. Then model is supervised to output single-class ``B-I-O'' result for ``Color'', given the scores and features.  

Formally, refer the prototype or textual embeddings of a selected attribute $a$ as $P^a_{min}\in \mathbb{R}^{1\times D}$, and each position $j$ of feature sequence encoded via $f$ for sample $i$ as $f_{ij}\in \mathbb{R}^{1\times D}$, $i$ is omitted for simplicity. Given an $a$, Feature Decoder decodes $f^a$ into ``B-I-O'' result for $a$. Our proposed guidance mechanism is denoted as:
\begin{equation}
	f_j^a=Concate(f_j, \frac{(P^a_{min})\ (f_j)^{\rm T}}{||P^a_{min}||\ ||f_j||}f_j).
	\label{eq8}
\end{equation}

Since our proposed Prototype-Guided Attribute Range Minimization (PAR) performs superiorly than other variants with an explicit design in the following section, it is embraced with TIR as a key component of our proposed Dynamic Range Minimization (DRAM) to achieve the best performance.

\section{Experiments}

In this section, extensive experiments are conducted on two popular multi-modal e-commerce attribute value extraction benchmarks, i.e., MEPAVE \cite{zhu2020multimodal} and MAE \cite{logan2017multimodal}, to evaluate our proposed ULS-DRAM method. Ablation study validates the effectiveness of ULS and two components TIR and PAR of DRAM. Furthermore, we also report the state-of-the-art comparison on both benchmarks.

\subsection{Datasets}

MEPAVE \cite{zhu2020multimodal}, named as Multimodal E-commerce Product Attribute Value Extraction dataset, collects both textual product descriptions and product images from a main-stream Chinese e-commerce platform JD.com. Given a sentence, the position of values mentioned in the sentence as well as the corresponding attributes are annotated together. 87,194 text-image instances with 26 types of product attributes are obtained from categories such as clothes, shoes, luggage, and bags. And the distribution of attributes among categories varies a lot. 71,194 instances are split into training set, 8000 for validation and testing set each. F1-score is the metric of performance on both attribute classification and attribute value extraction task.

MAE \cite{logan2017multimodal} is an English E-commerce Multimodal Attribute Extraction dataset. Categories are of much wider diversity including electronic products, jewelry, clothing, vehicles, and real estate. A textual product description, a collection of images and attribute-value pairs are collected in each record. 2.2M product items are split into training, validation and testing set with 80\%-10\%-10\% ratio, with 2k attribute classes annotated. Following \cite{zhu2020multimodal}, to avoid the attribute values which do not present in textual descriptions, MAE-text subset is constructed and denoted as MAE if not specially mentioned. Accuracy is adopted as the metric of attribute value extraction performance.

\subsection{Implementation Details}

Our proposed method is implemented on the basis of BiLSTM+CRF \cite{zheng2018opentag} feature decoder and PyTorch framework \cite{paszke2019pytorch}. Log-Likelihood loss function of CRF \cite{lafferty2001conditional} supervises the attribute value extraction task, and an extra Binary-CrossEntory loss optimizes the multi-label classification task of PAR.  RoBERTa-small-chinese \cite{liu2019roberta},  i.e., a variant of BERT, is specialized for Chinese dataset MEPAVE \cite{zhu2020multimodal}, and BERT-base-uncased \cite{devlin2018bert} for English dataset MAE \cite{logan2017multimodal}. Meanwhile, ResNet-152 \cite{he2016deep} pretrained on ImageNet \cite{deng2009imagenet} is adopted to encode image features. Under our proposed ULS training, Adam \cite{kingma2014adam} is used for optimization with initial learning rate $5\times10^{-5}$ for fine-tuning ResNet and $1\times10^{-4}$ for training other pertained or task-specific parameters. For MEPAVE, our model is trained and tested on one Nvidia V100 GPU with batch size 70. For MAE, it is trained with total batch size 560 on 8 GPUs and tested with 1000 on one GPU. Input images are resized to $224\times224$ following the settings of ImageNet dataset \cite{deng2009imagenet}.  We set $S_{thr}$ of proposed attribute range minimization to 0.5 and 0.4 respectively for MEPAVE and MAE via experimental trials.

\begin{table}[t]
	\begin{minipage}[b]{0.37\textwidth}
		\centering
		\renewcommand\thetable{1}
		\caption{Ablation study on our proposed ULS and TIR.}
		\begin{tabular}{c|c|c}
			\toprule[1.2pt]
			Scheme   & Method  & TAG-F1   \\
			\midrule[0.8pt]
			Fixed & M-JAVE\cite{zhu2020multimodal}  & 87.17 \\ 
			\midrule[0.8pt]
			\multirow{2}*{ULS-L}  &  BERT\cite{devlin2018bert} &   94.88  \\
			& RoBERTa\cite{liu2019roberta}	& 95.21	\\  
			\midrule[0.8pt]
			\multirow{4}*{ULS-VL}  
			& M-JAVE\cite{zhu2020multimodal} & 95.40 \\
			\cmidrule[0.5pt]{2-3}
			& Vanilla &  95.05 	   \\
			& Self-Attn & 94.89 \\  
			& TIR & {\bf 96.35} \\  
			\bottomrule[1.2pt]
		\end{tabular}
		\vspace{3mm}
		\label{table1}
	\end{minipage}
	\begin{minipage}[b]{0.60\textwidth}
		\centering
		\renewcommand\thetable{3}
		\caption{Comparison among variants of proposed Attribute Range Minimization (AR), including Prototype-AR, DyNet-AR and BERT-AR.}
		\begin{tabular}{c|cccc|cc}
			\toprule[1.2pt]
			Method & TIR & DAR & BAR & PAR & CLS-F1 & TAG-F1 \\
			\midrule[0.8pt]
			ULS-VL & & & & & - & 95.05 \\
			+TIR & \checkmark & & & & - & 96.35 \\
			\midrule[0.8pt]
			+DyN-AR & \checkmark & \checkmark & & & 97.52 & 96.07 \\
			+BERT-AR& \checkmark & & \checkmark & & 97.54 & 96.62 \\
			+Proto-AR & \checkmark & & & \checkmark & {\bf 97.65} & {\bf 96.86} \\		
			\bottomrule[1.2pt]
		\end{tabular}
		
		\vspace{6mm}
		\label{table3}
	\end{minipage}
	\begin{minipage}[b]{0.67\textwidth}
		\centering
		\renewcommand\thetable{2}
		\caption{Overall ablation study on key components of proposed ULS-DRAM, including PAR and TIR.}
		\begin{tabular}{c|c|c|c|c|c|c}
			\toprule[1.2pt]
			Method & Precision & Gain & Recall & Gain & TAG-F1 & Gain \\
			\midrule[0.8pt]
			ULS-VL & 94.04 & - & 96.07 & - & 95.05 & - \\
			+PAR & 94.99 & +0.95 & 96.71 & +0.64 & 95.84 & +0.79 \\
			+TIR & 95.21 & +1.17 &  97.52 & +1.45 & 96.35 & +1.30 \\
			+TIR+PAR & {\bf 95.89} & +1.85 & {\bf 97.86} & +1.79 & {\bf 96.86} & +1.81 \\
			\bottomrule[1.2pt]
		\end{tabular}
		\label{table2}
	\end{minipage}
	\hspace{0.05em}
	\begin{minipage}[b]{0.31\textwidth}
		\centering
		\renewcommand\thetable{5}
		\caption{Comparison with state-of-the-arts on MAE.}
		\begin{tabular}{c|c}
			\toprule [1.2pt]
			Method & Accuracy \\ 
			\midrule[0.8pt]
			MAE-model\cite{logan2017multimodal} & 72.96 \\
			M-JAVE\cite{zhu2020multimodal} & 75.01 \\
			ULS-DRAM$\dagger$ & {\bf 79.20} \\
			\bottomrule[1.2pt]
		\end{tabular}
		\label{table5}
		\vspace{2mm}
	\end{minipage}
\end{table}

\subsection{Ablation Study}

The ablation study is firstly performed on MEPAVE dataset. For attribute value extraction task in a  sequence tagging manner, F1-score, i.e., harmonic mean of Precision and Recall, is widely used to measure the comprehensive performance, denoted as TAG-F1. An extra CLS-F1 implies the attribute classification performance for minimizing the attribute range of input samples.
If not mentioned, percentage symbol (\%) is omitted for simplicity.

\let\thefootnote\relax\footnotetext{$\dagger$ Since the details of how to compute the ``Accuracy'' for sequence tagging methods are unavailable in \cite{zhu2020multimodal}, we have contacted the authors and adopted the following computation formula: Accuracy = num(correctly\_predicted\_attribute\_values) / num(total\_attribute\_values)}

Table \ref{table1} illustrates the ablation study on proposed Unified Learning Scheme (ULS) and Text-Guided Information Range Minimization (TIR). For ULS-L, the model is jointly trained with simple architecture consisting of pretrained linguistic model and single-modal textual input as shown in Figure \ref{figure3}, but still noticeably out-performs specially designed multi-modal  M-JAVE \cite{zhu2020multimodal} with fixed pretrained parameters. BERT-base-chinese \cite{devlin2018bert} and RoBERTa-small-chinese \cite{liu2019roberta} are used as candidate pretrained models, since BERT-small trained on Chinese corpus is unavailable. Due to larger size of parameters, in our opinion, BERT version of ULS-L model tends to over-fit on training subset. Therefore, RoBERTa version generalizes ideally and is chosen as our basic model.

For ULS-VL, multi-modal data with text and image are embraced to further jointly train not only pretrained linguistic model \cite{devlin2018bert} but also visual one \cite{he2016deep}. Referred as ``Vanilla'', ULS-VL adopts extra visual input V with textual one T as ``$Concate({\rm T}, {\rm V})$''. Its self-attention $QK^{\rm T}/\sqrt{D'}$ in BERT encoder can be divided into 4 parts: T$\rightarrow$T, V$\rightarrow$V, T$\rightarrow$V, V$\rightarrow$T. So \textit{all the possible guidances of attention are achieved}. Unfortunately, language-irrelated distractors ruin the model performance via fully guidance, i.e., TAG-F1 drops from 95.21 to 95.05. For further evaluation, an extra self-attention layer (``Self-Attn'') is added between BERT and input, then the TAG-F1 becomes even worse, only 94.89.

\setlength\intextsep{-8pt}
\begin{wraptable}[16]{r}{16.5em}
	\renewcommand\thetable{4}
	\caption{Comparison with state-of-the-arts on MEPAVE. ``*'' means extra data.}
	\vspace{0.2em}
	\begin{tabular}{c|cc}
		\toprule[1.2pt]
		Method & CLS-F1 & TAG-F1 \\
		\midrule[0.8pt]
		SUOpenTag \cite{xu2019scaling} & - & 77.12 \\
		RNN-LSTM \cite{hakkani2016multi} & 85.76 & 82.92 \\
		Attn-BiRNN \cite{liu16c_interspeech} & 86.10 & 83.28 \\
		Slot-Gated \cite{goo2018slot} & 86.70 & 83.35 \\
		Joint-BERT \cite{chen2019bert} & 86.93 & 83.73 \\
		JAVE-LSTM \cite{zhu2020multimodal} & 87.88 & 84.09 \\
		JAVE-BERT \cite{zhu2020multimodal} & 87.98 & 84.78 \\
		K-PLUG* \cite{xu2021k} & - & 95.97 \\
		\midrule
		M-JAVE-LSTM \cite{zhu2020multimodal} & 90.19 & 86.41 \\
		M-JAVE-BERT \cite{zhu2020multimodal} & 90.69 & 87.17 \\
		ULS-DRAM & {\bf 97.65} & {\bf 96.86} \\
		\bottomrule[1.2pt]
	\end{tabular}
	\label{table4}
\end{wraptable} 

Instead, our proposed TIR dynamically selects the descriptive parts of each modality into minimized information range. Wth text guidance T$\rightarrow$T and T$\rightarrow$V to focus on language-related information, TAG-F1 increases to 96.35. Besides, M-JAVE \cite{zhu2020multimodal} is fine-tuned with the settings of our ULS-VL and performs better than original fixed scheme. TIR with a strong linguistic encoder also out-performs M-JAVE with a naïve late fusion. In a word, the results prove that ULS effectively combines knowledge from pretraining and down-stream task, and TIR also suppresses the distractors of each modality via a minimized range to better construct a common semantic space by a strong encoder.

Table \ref{table2} further shows the performance gain by the key components of our proposed ULS-DRAM. Both ULS-VL+PAR and ULS-VL+TIR+PAR (i.e., ULS-DRAM) obtain a higher Precision gain than Recall, which significantly shows that our proposed PAR successfully decreases the false positives by attribute range minimization. And the overall TAG-F1 gains of both TIR and PAR are also prominent. In conclusion, based on ULS-VL, our proposed DRAM wins higher F1-score to further boost the attribute value extraction.

\subsection{Comparison among Variants of Proposed Attribute Range Minimization}

Then we evaluate the effectiveness of different policies for proposed Attribute Range Minimization, based on our ULS-VL model. Table \ref{table3} shows that DyNet Guided Policy leads to a slightly performance loss, which might be caused by high instability of dynamically applying parameters. BERT Guided Policy ranks the $2^{\rm nd}$ with more pretrained parameters than Prototype one. We think it might overfit while encoding the attribute words. Finally, with a compacted but powerful design, Prototype Guided Policy performs best on both CLS-F1 and TAG-F1. 

Consequently, Prototype-Guided Attribute Range Minimization, denoted as PAR, is selected as a fundamental component of our proposed DRAM.

\subsection{Comparison with the State-of-the-arts}

For MEPAVE, we compare our ULS-DRAM with not only single-modal but also multi-modal competitors, including SUOpenTag \cite{xu2019scaling}, RNN-LSTM \cite{hakkani2016multi}, Attn-BiRNN \cite{liu16c_interspeech}, Slot-Gated \cite{goo2018slot}, Joint-BERT \cite{chen2019bert}, K-PLUG \cite{xu2021k} (\textit{the model of which is pretrained with extra 25M data and further fine-tuned in a down-stream task, marked with ``*''}), single-modal JAVE \cite{zhu2020multimodal} and multi-modal M-JAVE \cite{zhu2020multimodal} based on LSTM \cite{hochreiter1997long} or BERT \cite{devlin2018bert}. SUOpenTag performs attribute extraction with already given attributes, thus the CLS-F1 is unavailable. Both Joint-BERT and SUOpenTag jointly fine-tune the pretrained BERT model \cite{devlin2018bert}, while the others only use pretrained word embeddings or fixed pretrained parameters. In Table \ref{table4}, our method achieves 97.65 CLS-F1 and 96.86 TAG-F1 as a significant performance gain via proposed ULS and DRAM, without any extra data.

For MAE, state-of-the-art methods including MAE-model \cite{logan2017multimodal} and M-JAVE \cite{zhu2020multimodal} are compared under multi-modal attribute value extraction task. Note that MAE-model is a generative model based on fixed pretrained parameters and given attribute inputs. While our ULS-DRAM and M-JAVE apply more stable sequence tagging and need no given attributes. Table \ref{table5} shows that our method achieves higher accuracy and improves the attribute value extraction on such a large-scale multi-modal dataset.

In conclusion, our ULS-DRAM has performed as a new state-of-the-art on both benchmarks, which sufficiently validates its capability of boosting attribute value extraction via proposed Unified Learning Scheme and Dynamic Range Minimization.

\section{Conclusion}

In this paper, we proposed a novel approach ULS-DRAM for boosting the multi-modal e-commerce attribute value extraction, which consists of two key components: ULS efficiently combines the common knowledge from up-stream pretraining and task-specific one from down-stream learning; DRAM minimizes the range of not only multi-modal information but also attributes with TIR and PAR.  Following the intuition that only language-related parts of each modality contribute to the extraction of textual values, TIR dynamically extracts the descriptive parts of vision and language, then encodes them into a common embedding space with a powerful pretrained linguistic model. PAR dynamically determines the range of attributes and selects the learnable prototypes to guide the prediction of the chosen attributes. We further investigated different policies for attribute range minimization: DyNet, BERT and Prototype Guided Policy, and the best-performing Prototype one is chosen. Complementing the proposed ULS and DRAM, a novel powerful model is then obtained and achieves new state-of-the-art performances on two challenging benchmarks MEPAVE and MAE. 

\appendix
	
\section{Appendix}

%\vspace{-1em}
\begin{figure}[t]
	\centering
	\includegraphics[width=1\textwidth]{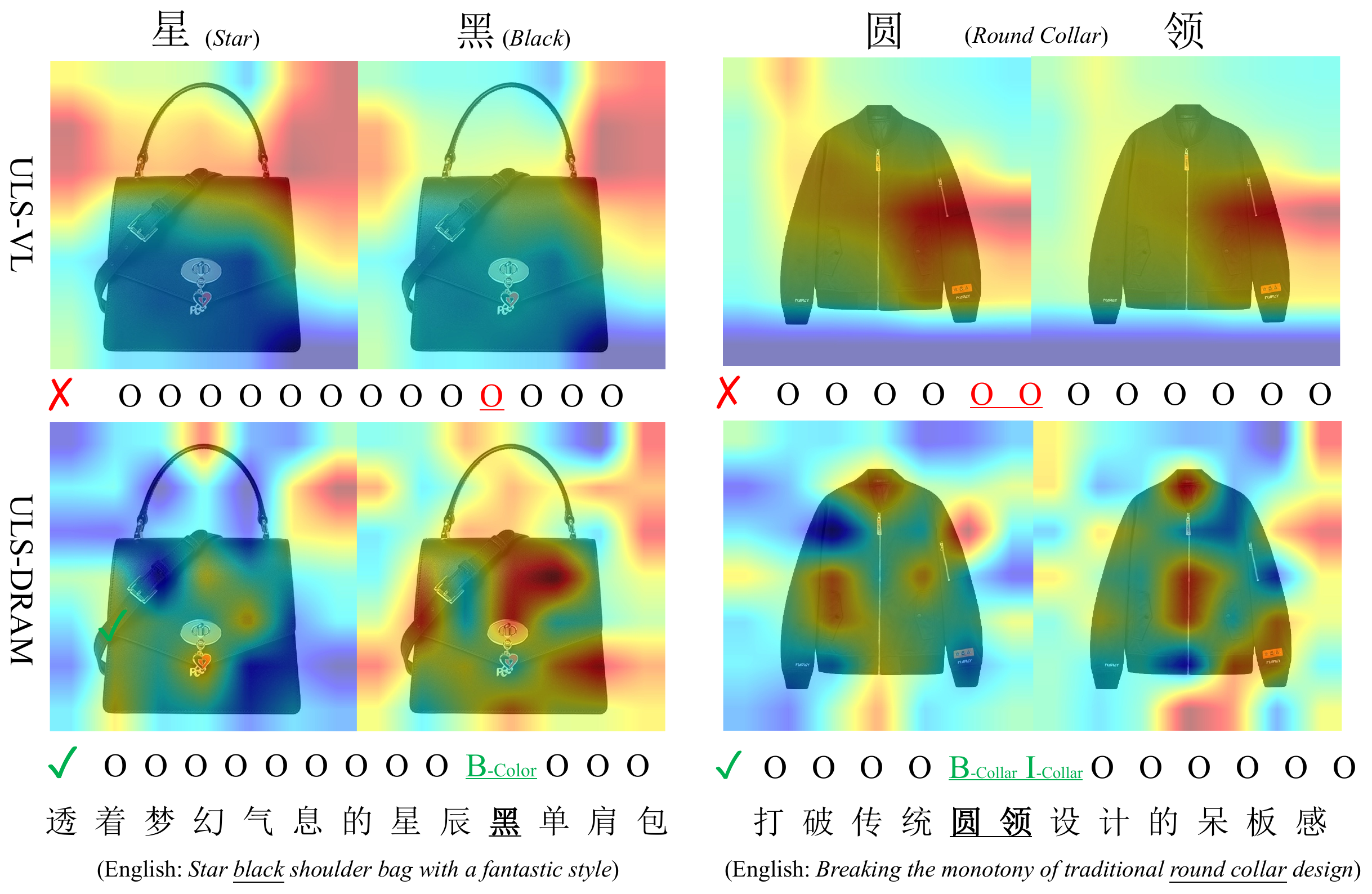}
	\caption{
		Qualitative results of proposed ULS-VL and ULS-DRAM on missing attribute value extractions. The $1^{\rm st}$ and $2^{\rm nd}$ rows of \underline{both heatmaps and ``B-I-O'' sequences} in each sample are generated by proposed ULS-VL and ULS-DRAM respectively. Columns of heatmaps are queried by corresponding word on the top to dynamically extract the descriptive parts of image regions. \textcolor{green}{\textbf{Green}} are right sequence predictions of attribute values while \textcolor{red}{\textbf{Red}} are missing ones. Heatmaps are generated from text-guided cross-attention in proposed TIR for ULS-DRAM, or from image and text-guided $1^{\rm st}$ self-attention encode layer of pretrained BERT\cite{devlin2018bert} for ULS-VL. The warmer the higher weight for selecting certain image region.}
	\label{figure6}
\end{figure}

\subsection{Qualitative Analysis}

In this section, further qualitative results on the popular benchmark MEPAVE\cite{zhu2020multimodal} between our proposed \textbf{ULS-DRAM} and fundamental method  ULS-VL are provided, in order to better understand the effectiveness of proposed \textbf{D}ynamic \textbf{RA}nge \textbf{M}inimization (\textbf{DRAM}). Not only predicted ``B-I-O'' sequences to extract attribute values, but also the attention heatmaps of each textual word to query the descriptive regions of image are shown for each sample, following \cite{zhu2020multimodal,huang2020pixel,DBLP:conf/icml/KimSK21}.

\begin{figure}[t]
	\centering
	\includegraphics[width=1\textwidth]{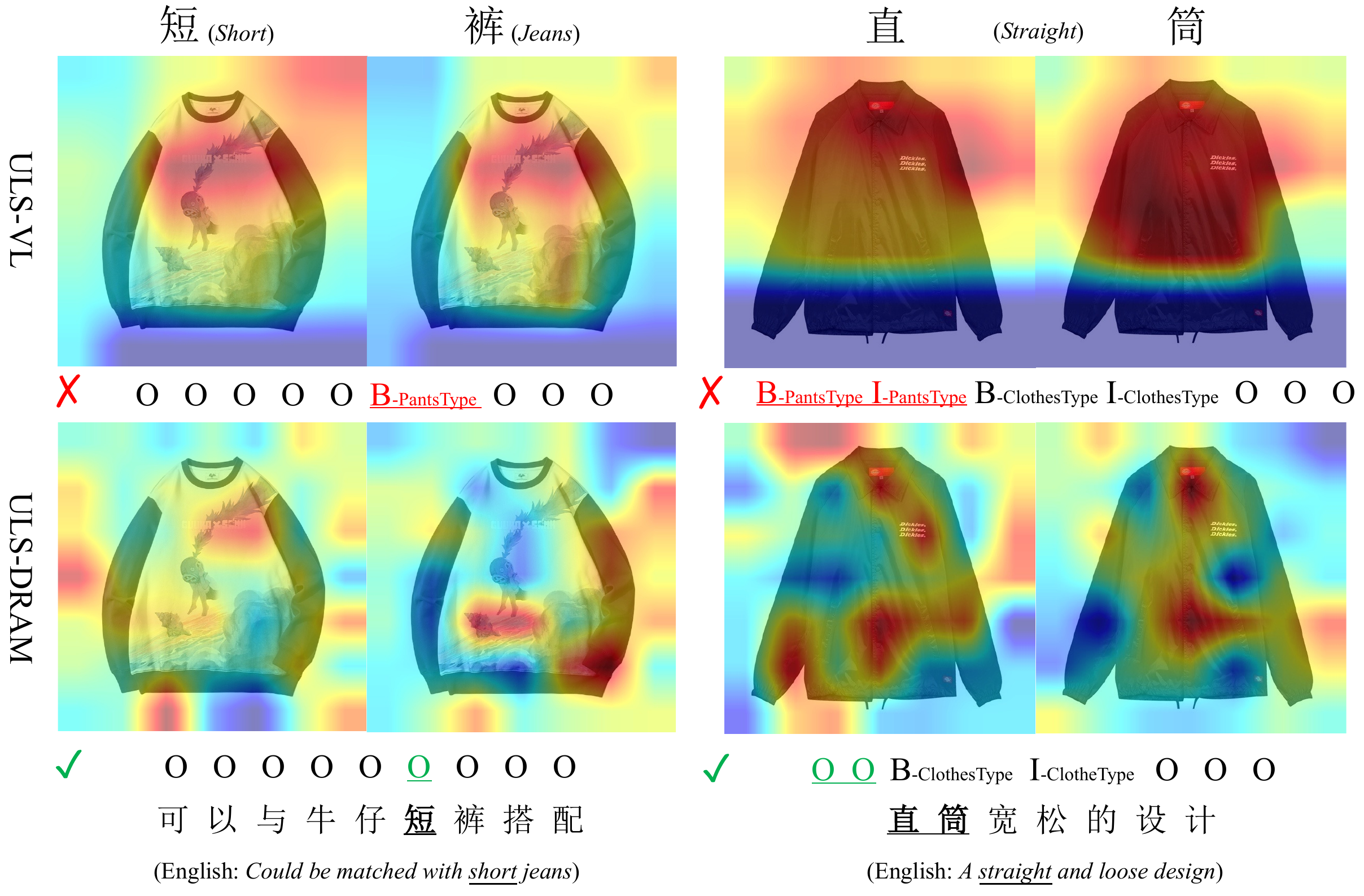}
	\caption{Qualitative results of proposed ULS-VL and ULS-DRAM on wrong attribute value extractions. The $1^{\rm st}$ and $2^{\rm nd}$ rows of \underline{both heatmaps and ``B-I-O'' sequences} in each sample are generated by proposed ULS-VL and ULS-DRAM respectively. Columns of heatmaps are queried by corresponding word on the top to dynamically extract the descriptive parts of image regions. \textcolor{green}{\textbf{Green}} are right sequence predictions of attribute values while \textcolor{red}{\textbf{Red}} are wrong ones. Heatmaps are generated following the settings of Figure \ref{figure6}. The warmer the higher weight for selecting certain image region.	}
	\label{figure7}
\end{figure}

\subsubsection{Comparison on Missing Extractions}

As is shown in Figure \ref{figure6}, missing attribute values in the $1^{\rm st}$  row of ``B-I-O'' sequences are highly correlated with the contextual information, from not only the textual words in neighborhood but also the language-related image regions, which requires an extractor to firstly dynamically select language-related visual clues, secondly encode them with textual ones into a common linguistic space, and finally make decision to extract attribute values based on semantic symbols in this space. 

Left part of Figure \ref{figure6} is an example that ``\begin{CJK}{UTF8}{gbsn}黑\end{CJK}'' (\textit{black}) inside ``\begin{CJK}{UTF8}{gbsn}星辰黑\end{CJK}'' (\textit{star black}) should be extracted as a color attribute value. However, ULS-VL made an empty prediction in the $1^{\rm st}$  row of ``B-I-O'' sequences with a ignorance of this value. More specifically, from the $1^{\rm st}$  row of heatmaps for word ``\begin{CJK}{UTF8}{gbsn}星\end{CJK}'' (\textit{star}) and ``\begin{CJK}{UTF8}{gbsn}黑\end{CJK}'' (\textit{black}), we can observe that lower attention is paid on the black-color surface of shoulder bag in this image. Worse still, similar heatmaps for words with different meanings represent that few of  language-related information is selected from input image, thus a language-agnostic  one is shared across the words. 

From the perspective of the model design, due to self-attention under the guidance of queries from both visual and linguistic input, $1^{\rm st}$ layer of pretrained BERT\cite{devlin2018bert} model in ULS-VL hardly strikes a balance between the guidance of two modalities. Hence, it is difficult for it to select the minimized range of language-related information, model a common linguistic space with semantic symbols, and then make decision on whether a textual phrase is a attribute value. 

\begin{figure}[t]
	\centering
	\includegraphics[width=1\textwidth]{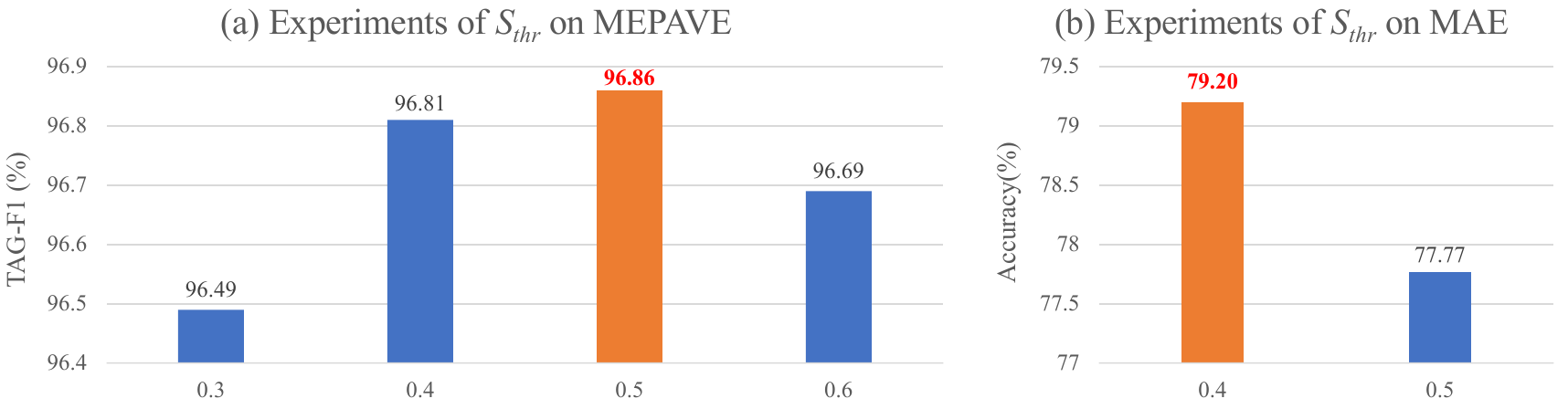}
	\caption{Experiments of $S_{thr}$ on different benchmarks. As the threshold to filter the attribute with higher score in multi-label classification of proposed Prototype-Guided Attribute Range Minimization (PAR), $S_{thr}$ varies due to the specific attribute distribution of each benchmarks. \textcolor{red}{\textbf{Red}} are the best performances with selected $S_{thr}$ on MEPAVE and MAE, i.e., $0.5$ and $0.4$ respectively.}
	\label{figure8}
\end{figure}

Differently, in the 2nd row on the left of Figure \ref{figure6}, with the help of proposed Text-Guided Information Range Minimization, ULS-DRAM generates two distinct heatmaps for two different words. For ``\begin{CJK}{UTF8}{gbsn}星\end{CJK}'' (\textit{star}), the extractor pays attention on the contextual image regions, because there are not any stars actually in this image (\textit{p.s., ``star'' is only a modifier for the black color}). For ``\begin{CJK}{UTF8}{gbsn}黑\end{CJK}'' (\textit{black}), most attention is applied on the black surface of shoulder bag in the image. Consequently, language-related information from image are dynamically extracted in a minimized range by our proposed ULS-DRAM method to predicts correct sequences in the $2^{\rm nd}$  row.

Another example on the right of Figure \ref{figure6} shows a word phrase ``\begin{CJK}{UTF8}{gbsn}圆领\end{CJK}'' (\textit{round collar}) should be treated as a attribute value. Still, ULS-VL generates similar heatmaps of both words with lower score on the collar in the image, which leads to the missing extraction based on visual information unrelated to language. Dissimilarly, ULS-DRAM not only generates heatmaps focusing more on the collar region, but also keeps the consistency across the two word ``\begin{CJK}{UTF8}{gbsn}圆\end{CJK}'' (\textit{round}) and ``\begin{CJK}{UTF8}{gbsn}领\end{CJK}'' (\textit{collar}), so that both of the two words of the whole word phrase benefit from the same language-related visual clues in this image and correct prediction are made. 

As a summary, it is highlighted in Figure \ref{figure6} that our proposed ULS-DRAM minimizes the information range with the guidance of language, and takes full advantage of the semantic symbols from multiple modalities in a common space modeled by pretrained BERT\cite{devlin2018bert} encoder, rather than directly concatenates visual and textual sequences and feed them into pretrained encoder as ULS-VL does.

\subsubsection{Comparison on Wrong Extractions}

In Figure \ref{figure7}, some attribute values are extracted by mistake, which depends on the capability of attribute range minimization for an extractor to eliminate the non-essential predictions. Because of the diversity among products, there are different attribute ranges of each item, e.g., the attributes of pants are not totally shared with those of clothes. Maintaining the consistency between attribute distribution and modeling, an extractor will performs better within a dynamically minimized attribute range.

As is illustrated in $1^{\rm st}$ row of ``B-I-O'' sequence on the left of Figure \ref{figure7}, an attribute value for pants type ``\begin{CJK}{UTF8}{gbsn}短\end{CJK}'' (\textit{short}) is mistakenly extracted for a shirt in the image. Considering attribute ranges of all the products as a maximal one, ULS-VL outputs the scores of all the attributes at each position of ``B-I-O" sequence. Thus it is possible for attributes irrelevant with shirt to be predicted, especially when inconsistent shirt regions are extracted for both ``\begin{CJK}{UTF8}{gbsn}短\end{CJK}'' (\textit{short}) and ``\begin{CJK}{UTF8}{gbsn}裤\end{CJK}'' (\textit{jeans}) in the heatmaps of $1^{\rm st}$ row. 

Cooperating with Text-Guided Information Range Minimization (TIR), our proposed ULS-DRAM selects minimal shirt regions for ``\begin{CJK}{UTF8}{gbsn}短\end{CJK}'' (\textit{short}) and lower part of shirt, as it is more likely to appear a jeans, for ``\begin{CJK}{UTF8}{gbsn}裤\end{CJK}'' (\textit{jeans}). Further improved by our proposed Prototype-Guided Attribute Range Minimization (PAR), ULS-DRAM successfully avoids the interruption of shirt information from images and keep the attribute range with shirt-related minimized one. 

For a two-word phrase ``\begin{CJK}{UTF8}{gbsn}直筒\end{CJK}'' (\textit{straight}) in right part of Figure \ref{figure7}, ULS-VL pays its most attention on the jacket in the image, but still ignores the specialized attribute range for the jacket. On the contrary, our proposed ULS-DRAM captures the ``straight'' part of jacket via TIR, such as the sleeve in bottom left heatmap and lappet with zipper in bottom right heatmap. With the assistance of PAR, it succeeds in keeping the attribute range minimized for jacket, with only ``ClothesType'' attribute value ``\begin{CJK}{UTF8}{gbsn}宽松\end{CJK}'' (\textit{loose}) extracted.

In conclusion, the qualitative results reveal that complementing TIR and PAR, our proposed ULS-DRAM dynamically selects the language-related information and encodes them into a common space for decision making, minimizes the attribute range according to input sample to eliminate wrong predictions, and thus improves the performance of multi-modal attribute value extraction.

\subsection{Selection of $S_{thr}$  on Different Benchmarks}

Considering the diversified attribute distribution of different benchmarks, multiple values of the hyper-parameter $S_{thr}$ are experimented in Figure \ref{figure8}, which are used for the threshold to filter high-score attributes into minimized ranges of proposed PAR. 

For the relatively compacted and balanced 26 attributes in MEPAVE \cite{zhu2020multimodal}, higher threshold of  $S_{thr}$ encourages the attribute range to converge into a smaller minimized range. Therefore, as is shown in Figure \ref{figure8} (a), $S_{thr}=0.5$ achieves best TAG-F1 score over the other candidates. For the distracted and long-tailed 2k attributes in MAE \cite{logan2017multimodal}, lower $S_{thr}$ ensures that all the potential attributes are comprised in the minimized attribute range, preventing the absence of attributes with lower frequency. Hence, $S_{thr}=0.4$ improves the performance of $0.5$ with a significant gain in Figure \ref{figure8} (b). Due to the constraint memory budget of GPUs, lower values of $S_{thr}$ are not experimented. Note that it is just a \textbf{priori knowledge that $S_{thr}$ is selected according to data distribution}, and even default 0.5 achieves state-of-the-art performance in Tables 4 and 5.

\subsection{Computation of the Accuracy for MAE}

Since the details of the ``Accuracy'' computation on MAE for sequence tagging methods including \cite{zhu2020multimodal} and our proposed ULS-DRAM are unavailable, having contacted the authors of \cite{zhu2020multimodal}, we adopt the following computation formula:
\setcounter{equation}{8}
\begin{equation}
	Accuracy = \frac{|V_{correct}|}{|V_{total}|}.
	\label{eq9}
\end{equation}

$V_{correct}$ is the set of correctly predicted attribute-value pairs and $V_{total}$ is all the annotated pairs. An attribute pair is obtained via post-processing of the ``B-I-O'' sequence along with the original textual input. Here is a example:
\begin{quote}
	Input text: ``blue jeans match blue spirits'' \\
	Annotation: ``Color: [blue]''
	\\\\
	Prediction (1): \\``B-Color O O B-Color O'' or ``B-Color O O O O''.\\
	Pairs: ``Color: [blue]'', Correct.
	\\
	Prediction (2): \\``B-Color O O O B-Color''. \\
	Pairs: ``Color: [blue, spirits]'', Incorrect.
	\\
	Prediction (3): \\``O O O O O'' or ``O O O B-Color I-Color''. \\
	Pairs: ``Color: []'' or ``[blue spirits]'', Incorrect.
	
\end{quote}

Due to the original MAE dataset is annotated for generative MAE-model, the annotations are not sequences but attribute-value pairs for directly compute the top-k hit rates of the generated candidate results, e.g., $Hit@1$ or $Hit@5$. The attributes of the corresponding ``B-I'' sequence are marked in the append names of each ``B'' and ``I'', e.g., ``B-Color'' or ``I-Pattern''.  And the values can be extracted from the same positions of ``B''s and ``I''s in original text sequences. 

To determine the correctness of predicted ``B-I-O'' sequence, they are post-processed into attribute-value pairs. In $1^{\rm st}$ sequence in Prediction (1), two values ``blue'' are extracted but they are merged into one, as the $2^{\rm nd}$ sequence. Although annotated value ``blue'' are predicted in Prediction (2), an extra ``spirits'' is extracted and leads to incorrectness. As long as none of the annotated values are predicted, as in Prediction (3), the results are incorrect. In a word, the predicted attribute-value pairs should be absolutely the same with  corresponding annotated one.
\begin{quote}
	Prediction (4): \\``B-Color B-Type O B-Color O''.\\
	Pairs: ``Color: [blue], Type: [jeans]'', Correct.
\end{quote}

Note that the \textbf{extra attributes} beyond annotation are not considered during determination of correctness, as shown in Prediction (4) above. Maybe some argue that ``false positives'' will be ignored, but the actual ``false positives'' are similar to ``spirits'' or ``blue spirits'' in Prediction (2) and (3) within the range of annotated attributes. 

Once the correctness of each predicted attribute-value pair is available, the $Accuracy$ in Eq. \ref{eq9} is computed simply via counting the number of correctly predicted pairs and totally annotated ones. The default maximum of $Accuracy$ is $100\%$, which stands for all the predicted pairs are absolutely the same with annotated one.

\subsection{Discussions on Limitations}

Our proposed multi-modal attribute value extraction method ULS-DRAM is designed for e-commerce scenarios, where data distributions of both visual and textual modality are featured with domain-specific characteristics. Hence, there are some limitations in its application.

On the one hand, during training and inference time, most images consist of these fundamental parts: (1) a singleton of product item in a closer-look perspective, (2) a clean background (\textit{sometimes filled with pure colors}) with minimal disturbance to product, and (3) additional commercial elements such as brand names or trade marks. 

Therefore, \textbf{multi-modal attribute value extractors are seldomly affected by processing too much background regions} with scarce noise and low information density. On the contrary, some ignorance of descriptive parts leads to incorrect predictions. As is shown in Figure \ref{figure6} and \ref{figure7}, even though our ULS-DRAM focuses on a few regions of background, descriptive parts and their contexts , e.g. the surfaces of a bag or the collar of a jacket, are paid attention correctly to assist avoiding mistakes. Consequently, our proposed method might be not capable enough of processing images with more noisy background, such as some photos on social media platforms or news websites.

On the other hand, \textbf{the textual data in e-commerce scenarios tends to be brief and descriptive},  without any story telling or technical terms. For instance, it is rather difficult for our method to understand a drama and extract the attribute values of specific characters, or read long scientific papers to summarize them with the keywords as a series of attribute-value pairs. 

In conclusion, \textbf{the data distribution of e-commerce domain} is the major limitation of our proposed method, and it also provides the priori knowledge for us to obtain a specialized design and achieve better performances. 

% ---- Bibliography ----
%
% BibTeX users should specify bibliography style 'splncs04'.
% References will then be sorted and formatted in the correct style.
%
\bibliographystyle{splncs04}
\bibliography{egbib}

\begin{thebibliography}{10}
\providecommand{\url}[1]{\texttt{#1}}
\providecommand{\urlprefix}{URL }
\providecommand{\doi}[1]{https://doi.org/#1}

\bibitem{chen2019bert}
Chen, Q., Zhuo, Z., Wang, W.: Bert for joint intent classification and slot
  filling. arXiv preprint arXiv:1902.10909  (2019)

\bibitem{chen2020uniter}
Chen, Y.C., Li, L., Yu, L., El~Kholy, A., Ahmed, F., Gan, Z., Cheng, Y., Liu,
  J.: {UNITER}: Universal image-text representation learning. In: European
  conference on computer vision. pp. 104--120 (2020)

\bibitem{deng2009imagenet}
Deng, J., Dong, W., Socher, R., Li, L.J., Li, K., Fei-Fei, L.: Imagenet: A
  large-scale hierarchical image database. In: 2009 IEEE conference on computer
  vision and pattern recognition. pp. 248--255. {IEEE} (2009)

\bibitem{devlin2018bert}
Devlin, J., Chang, M.W., Lee, K., Toutanova, K.: {BERT:} pre-training of deep
  bidirectional transformers for language understanding. In: Proceedings of the
  2019 Conference of the North American Chapter of the Association for
  Computational Linguistics: Human Language Technologies, {NAACL-HLT}, Volume 1
  (Long and Short Papers). pp. 4171--4186 (2019)

\bibitem{dosovitskiy2020image}
Dosovitskiy, A., Beyer, L., Kolesnikov, A., Weissenborn, D., Zhai, X.,
  Unterthiner, T., Dehghani, M., Minderer, M., Heigold, G., Gelly, S., et~al.:
  An image is worth 16x16 words: Transformers for image recognition at scale.
  In: 9th International Conference on Learning Representations, {ICLR} 2021
  (2021)

\bibitem{gao2019product}
Gao, S., Ren, Z., Zhao, Y., Zhao, D., Yin, D., Yan, R.: Product-aware answer
  generation in e-commerce question-answering. In: Proceedings of the Twelfth
  ACM International Conference on Web Search and Data Mining. pp. 429--437
  (2019)

\bibitem{goo2018slot}
Goo, C.W., Gao, G., Hsu, Y.K., Huo, C.L., Chen, T.C., Hsu, K.W., Chen, Y.N.:
  Slot-gated modeling for joint slot filling and intent prediction. In:
  Proceedings of the 2018 Conference of the North American Chapter of the
  Association for Computational Linguistics: Human Language Technologies,
  Volume 2 (Short Papers). pp. 753--757 (2018)

\bibitem{hakkani2016multi}
Hakkani-T{\"u}r, D., T{\"u}r, G., Celikyilmaz, A., Chen, Y.N., Gao, J., Deng,
  L., Wang, Y.Y.: Multi-domain joint semantic frame parsing using
  bi-directional rnn-lstm. In: Interspeech. pp. 715--719 (2016)

\bibitem{he2016deep}
He, K., Zhang, X., Ren, S., Sun, J.: Deep residual learning for image
  recognition. In: Proceedings of the IEEE conference on computer vision and
  pattern recognition. pp. 770--778 (2016)

\bibitem{hochreiter1997long}
Hochreiter, S., Schmidhuber, J.: Long short-term memory. Neural computation
  \textbf{9}(8),  1735--1780 (1997)

\bibitem{hou2019explainable}
Hou, M., Wu, L., Chen, E., Li, Z., Zheng, V.W., Liu, Q.: Explainable fashion
  recommendation: {A} semantic attribute region guided approach. In:
  Proceedings of the Twenty-Eighth International Joint Conference on Artificial
  Intelligence, {IJCAI}. pp. 4681--4688 (2019)

\bibitem{huang2020pixel}
Huang, Z., Zeng, Z., Liu, B., Fu, D., Fu, J.: Pixel-bert: Aligning image pixels
  with text by deep multi-modal transformers. arXiv preprint arXiv:2004.00849
  (2020)

\bibitem{DBLP:conf/icml/KimSK21}
Kim, W., Son, B., Kim, I.: Vilt: Vision-and-language transformer without
  convolution or region supervision. In: Meila, M., Zhang, T. (eds.)
  Proceedings of the 38th International Conference on Machine Learning, {ICML}
  2021, 18-24 July 2021, Virtual Event. Proceedings of Machine Learning
  Research, vol.~139, pp. 5583--5594. {PMLR} (2021),
  \url{http://proceedings.mlr.press/v139/kim21k.html}

\bibitem{kingma2014adam}
Kingma, D.P., Ba, J.: Adam: {A} method for stochastic optimization. In: 3rd
  International Conference on Learning Representations, {ICLR} 2015, San Diego,
  CA, USA, May 7-9, 2015, Conference Track Proceedings (2015)

\bibitem{lafferty2001conditional}
Lafferty, J., McCallum, A., Pereira, F.C.: Conditional random fields:
  Probabilistic models for segmenting and labeling sequence data. In:
  Proceedings of the Eighteenth International Conference on Machine Learning
  {(ICML}. pp. 282--289 (2001)

\bibitem{lin2021pam}
Lin, R., He, X., Feng, J., Zalmout, N., Liang, Y., Xiong, L., Dong, X.L.:
  {PAM}: Understanding product images in cross product category attribute
  extraction. In: Proceedings of the 27th ACM SIGKDD Conference on Knowledge
  Discovery \& Data Mining. pp. 3262--3270 (2021)

\bibitem{liu16c_interspeech}
Liu, B., Lane, I.: {Attention-Based Recurrent Neural Network Models for Joint
  Intent Detection and Slot Filling}. In: Proc. Interspeech 2016. pp. 685--689
  (2016)

\bibitem{liu2019roberta}
Liu, Y., Ott, M., Goyal, N., Du, J., Joshi, M., Chen, D., Levy, O., Lewis, M.,
  Zettlemoyer, L., Stoyanov, V.: {RoBERTa}: A robustly optimized bert
  pretraining approach. arXiv preprint arXiv:1907.11692  (2019)

\bibitem{logan2017multimodal}
Logan~IV, R.L., Humeau, S., Singh, S.: Multimodal attribute extraction. In: 6th
  Workshop on Automated Knowledge Base Construction, AKBC@NIPS (2017)

\bibitem{paszke2019pytorch}
Paszke, A., Gross, S., Massa, F., Lerer, A., Bradbury, J., Chanan, G., Killeen,
  T., Lin, Z., Gimelshein, N., Antiga, L., et~al.: Pytorch: An imperative
  style, high-performance deep learning library. Advances in neural information
  processing systems  \textbf{32},  8026--8037 (2019)

\bibitem{shaw2018self}
Shaw, P., Uszkoreit, J., Vaswani, A.: Self-attention with relative position
  representations. In: Proceedings of the 2018 Conference of the North American
  Chapter of the Association for Computational Linguistics: Human Language
  Technologies, NAACL-HLT, Volume 2 (Short Papers). pp. 464--468 (2018)

\bibitem{su2019vl}
Su, W., Zhu, X., Cao, Y., Li, B., Lu, L., Wei, F., Dai, J.: {VL-BERT:}
  pre-training of generic visual-linguistic representations. In: 8th
  International Conference on Learning Representations, {ICLR} (2020)

\bibitem{sun2021rpbert}
Sun, L., Wang, J., Zhang, K., Su, Y., Weng, F.: {RpBERT}: A text-image relation
  propagation-based bert model for multimodal ner. In: Proceedings of the AAAI
  Conference on Artificial Intelligence. vol.~35, pp. 13860--13868 (2021)

\bibitem{wang2020learning}
Wang, Q., Yang, L., Kanagal, B., Sanghai, S., Sivakumar, D., Shu, B., Yu, Z.,
  Elsas, J.: Learning to extract attribute value from product via question
  answering: A multi-task approach. In: Proceedings of the 26th ACM SIGKDD
  International Conference on Knowledge Discovery \& Data Mining. pp. 47--55
  (2020)

\bibitem{xian2021ex3}
Xian, Y., Zhao, T., Li, J., Chan, J., Kan, A., Ma, J., Dong, X.L., Faloutsos,
  C., Karypis, G., Muthukrishnan, S., et~al.: Ex3: Explainable attribute-aware
  item-set recommendations. In: Fifteenth ACM Conference on Recommender
  Systems. pp. 484--494 (2021)

\bibitem{xu2019scaling}
Xu, H., Wang, W., Mao, X., Jiang, X., Lan, M.: Scaling up open tagging from
  tens to thousands: Comprehension empowered attribute value extraction from
  product title. In: Proceedings of the 57th Annual Meeting of the Association
  for Computational Linguistics. pp. 5214--5223 (2019)

\bibitem{xu2021k}
Xu, S., Li, H., Yuan, P., Wang, Y., Wu, Y., He, X., Liu, Y., Zhou, B.:
  {K-PLUG:} knowledge-injected pre-trained language model for natural language
  understanding and generation in e-commerce. In: Findings of the Association
  for Computational Linguistics: {EMNLP} 2021. pp. 1--17 (2021)

\bibitem{yan2021adatag}
Yan, J., Zalmout, N., Liang, Y., Grant, C., Ren, X., Dong, X.L.: Adatag:
  Multi-attribute value extraction from product profiles with adaptive
  decoding. In: Proceedings of the 59th Annual Meeting of the Association for
  Computational Linguistics and the 11th International Joint Conference on
  Natural Language Processing, {ACL/IJCNLP}. pp. 4694--4705 (2021)

\bibitem{zheng2018opentag}
Zheng, G., Mukherjee, S., Dong, X.L., Li, F.: {OpenTag}: Open attribute value
  extraction from product profiles. In: Proceedings of the 24th ACM SIGKDD
  International Conference on Knowledge Discovery \& Data Mining. pp.
  1049--1058 (2018)

\bibitem{zhu2020multimodal}
Zhu, T., Wang, Y., Li, H., Wu, Y., He, X., Zhou, B.: Multimodal joint attribute
  prediction and value extraction for e-commerce product. In: Proceedings of
  the 2020 Conference on Empirical Methods in Natural Language Processing,
  {EMNLP}. pp. 2129--2139 (2020)

\end{thebibliography}

\setcounter{figure}{5}
\setcounter{section}{5}

\end{document}